\documentclass[10pt,twocolumn,letterpaper]{article}
\usepackage{cvpr}
\usepackage{times}
\usepackage{epsfig}
\usepackage{graphicx}
\usepackage{amsmath}
\usepackage{subcaption}
\usepackage{amssymb}
\usepackage{multirow}
\usepackage{floatrow}
\usepackage{booktabs}
\usepackage{caption}
\usepackage{enumitem}
\usepackage{arydshln}
\usepackage{bm}
\usepackage{authblk}

\usepackage{makecell}
\usepackage{multicol}
\usepackage{bbm}
\usepackage{xcolor}
\usepackage{xspace}
\usepackage{float}
\usepackage{color}
\usepackage{colortbl}
\usepackage{lipsum}

\renewcommand{\Re}{\mathbb{R}}
\newcommand{\hy}{\hat{y}}
\newcommand{\hb}{\hat{b}}
\newcommand{\ha}{\hat{a}}
\newcommand{\hp}{\hat{p}}

\renewcommand{\Sigma}{\mathfrak{S}}

\newcommand{\indic}[1]{\mathds{1}_{\{#1\}}}

\newcommand{\bloss}[1]{{\cal L}_{\rm box}(#1)}
\newcommand{\aloss}[1]{{\cal L}_{\rm angle}(#1)}
\newcommand{\iouloss}[1]{{\cal L}_{\rm iou}(#1)}

\newcommand{\hloss}[1]{{\cal L}_{\rm Hungarian}(#1)}

\newcommand{\lmatch}[1]{{\cal L}_{\rm match}(#1)}

\def\eqref#1{equation~\ref{#1}}

\def\1{\bm{1}}

\DeclareMathAlphabet{\mathsfit}{\encodingdefault}{\sfdefault}{m}{sl}
\SetMathAlphabet{\mathsfit}{bold}{\encodingdefault}{\sfdefault}{bx}{n}

\DeclareMathOperator*{\argmin}{arg\,min}

\newcommand{\noobject}{\varnothing}

\newcolumntype{I}{!{\vrule width 3pt}}

\usepackage[ruled]{algorithm2e}             
\usepackage{array}

\newlength\savedwidth

\newlength\savewidth
\newcommand\shline{\noalign{\global\savewidth\arrayrulewidth
                           \global\arrayrulewidth 0.5pt}%
                  \hline
                  \noalign{\global\arrayrulewidth\savewidth}}

\def\eg{\emph{e.g.,}}

\def\ie{\emph{i.e.,}}

\usepackage[breaklinks=true,bookmarks=false]{hyperref}
\hypersetup{
colorlinks=true,
linkcolor=black
}

\def\varnothing{\emptyset}
\newcommand{\detr}{\textup{Trans}\textsc{\textbf{D}e\textbf{TR}}\xspace}
 
\renewcommand{\texttt}[1]{ $ {{\tt #1} } $}  
\usepackage{dsfont}

\usepackage[margin=4pt,font=small,labelfont=bf,labelsep=endash,tableposition=top]{caption}

\definecolor{mygray}{gray}{.92}

\cvprfinalcopy 

\def\cvprPaperID{5639} 

\begin{document}

\title{End-to-End Video Text Spotting with Transformer}

\author{
{\large
Weijia Wu$^1$},
{\large
Debing Zhang$^2$},
{\large
Ying Fu$^3$},
{\large
Chunhua Shen$^1$},
{\large
Hong Zhou$^1$$^\dagger$},\\
{\large
Yuanqiang Cai$^4$$^\dagger$},
{\large
Ping Luo$^5$}
}

\author{
{\large
Weijia Wu$^1$},
{\large
Yuanqiang Cai$^2$},
{\large
Chunhua Shen$^1$},
{\large
Debing Zhang$^3$},
{\large
Ying Fu$^4$},
{\large
Hong Zhou$^1$$^\dagger$},
{\large
Ping Luo$^5$}\\
{\large
$^1$Zhejiang University$ \qquad $}\\
{\large
$^2$Beijing University of Posts and Telecommunications $ \qquad $
}\\
{\large
$^3$Kuaishou Technology$ \qquad $}\\
{\large
$^4$Beijing Institute of Technology $ \qquad $
}\\
{\large
$^5$The University of Hong Kong 
}
}

\maketitle
\begin{abstract}
Recent video text spotting methods usually require
the three-staged pipeline, \ie{} detecting text in individual images, recognizing localized text, tracking text streams with post-processing to generate final results.
The previous methods typically follow the tracking-by-match paradigm and develop sophisticated pipelines, which is an not effective solution.
In this paper, rooted in Transformer sequence modeling, we propose a simple, yet effective end-to-end trainable video text \textbf{D}Etection, \textbf{T}racking, and \textbf{R}ecognition framework (\detr), which views the VTS task as a direct long-range temporal modeling problem.
\detr mainly includes two advantages: 
1) Different from the explicit match paradigm in the adjacent frame, the proposed \detr tracks and recognizes each text implicitly by the different query termed `text query' over long-range temporal sequence~(more than 7 frames).
2) \detr is the first end-to-end trainable video text spotting framework, which simultaneously addresses the three sub-tasks~(\eg{} text detection, tracking, recognition).
Extensive experiments on four video text datasets (\eg{} ICDAR2013 Video, ICDAR2015 Video) are conducted to demonstrate that \detr achieves the state-of-the-art performance with up to $11.0\%$ improvements on detection, tracking, and spotting tasks.
The code can be found at \href{https://github.com/weijiawu/TransDETR}{\color{blue}{$\tt github.com/weijiawu/TransDETR$}}.

\end{abstract}

\section{Introduction}
Text spotting is one of the fundamental tasks in computer vision.
In recent years, image-based text spotting has made extraordinary progress with a lot of great algorithms~\cite{zhou2017east,baek2019character,masktextspotter,psenet,liu2020abcnet,wu2020synthetic,wu2019textcohesion}, but much less effort was spent on video text spotting, which is the base of video-and-language tasks such as video retrieval~\cite{dong2021dual,NASA}, video caption~\cite{srivastava2015unsupervised} and driving navigation systems~\cite{reddy2020roadtext}.
Yin \textit{et al.}~\cite{yin2016text} firstly surveys and provides a detailed introduction for video text spotting task~(VTS), which requires algorithms to simultaneously detect, track and recognize each text instance in a video sequence. 
Compared with text spotting in static image~\cite{karatzas2015icdar}, video text spotting is more challenging. 
The quality of the video frames is generally worse than that of static images, due to motion blur, out of focus, and artifacts issues.
More importantly, the video text spotting task requires text spatio-temporal information, \ie{} the same text tracking trajectory(Tracking ID), which cannot be provided by image-based text spotting methods.

\begin{figure*}[t]
\begin{center}
\includegraphics[width=0.98\textwidth]{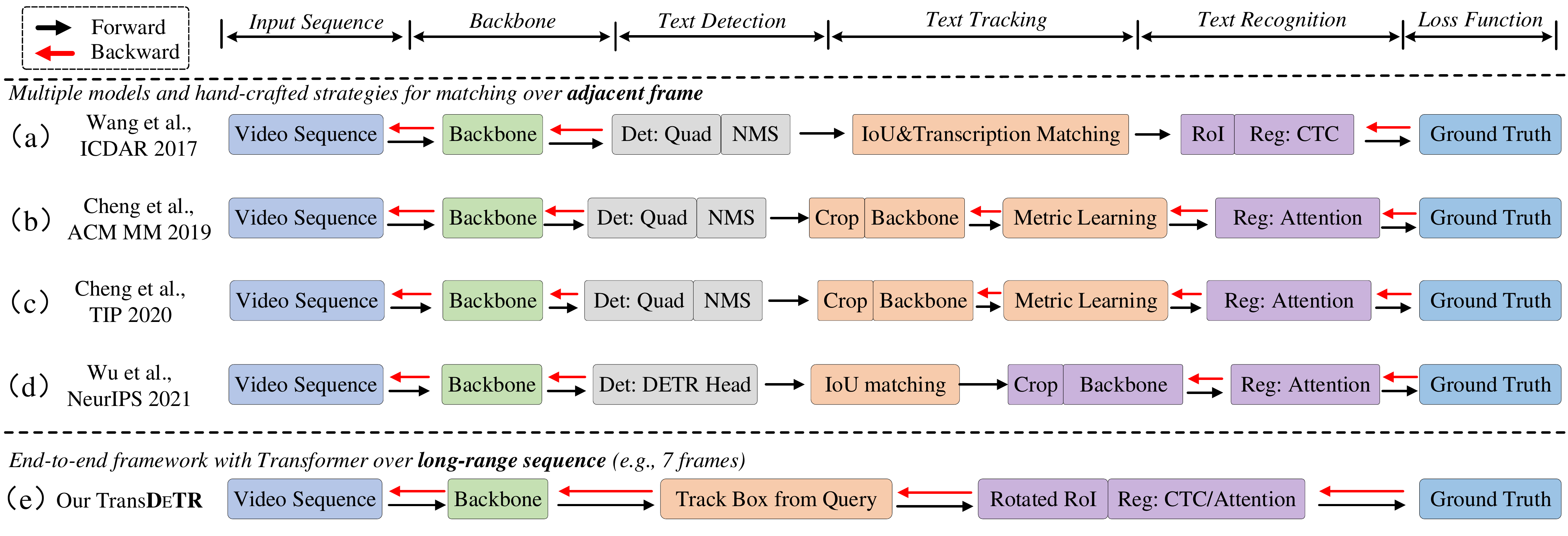}
\caption{\textbf{Comparisons of different video text spotting pipeline}. (a) Wang \textit{et al.}~\cite{wang2017end} tracks and recognizes text by transcription matching; (b-c) Cheng \textit{et al.}~\cite{cheng2019you,cheng2020free} tracks and recognizes text with metric learning; (d) Wu \textit{et al.}~\cite{wu2021bilingual} adopts IoU-based match to address the task; (e) Without complicated matching, hand-crafted components, an simpler, end-to-end pipeline with transformer is proposed in this work.}
\label{fig2}
\end{center}
\end{figure*}

Existing video text spotting methods typically develop sophisticated pipelines to tackle the task and view the video text spotting tasks as the text instances match problem in adjacent frames.
As shown in Figure~\ref{fig2} (a)-(e), the previous methods~\cite{wu2021bilingual,cheng2020free,wang2017end} usually use \textbf{multiple models} and \textbf{hand-crafted strategies} for the video text spotting task, which firstly tackles each frame in a video sequence, and then associates the similar text instances in the adjacent frames by various matching strategies(\ie{use their IoU, transcription, and feature}). 
For example, IoU-based methods~\cite{wu2021bilingual,rong2014scene} relies on image text detection models~\cite{psenet,zhou2017east} and complex human-designed matching rules, where
two text instances are associated while their IoU is higher than the given threshold. And one separate recognition model is used to recognize the text content.
Similarly, transcription-based~\cite{mita2001improvement,wang2017end} and Re-ID-based methods~\cite{cheng2019you,feng2021semantic} compute the text recognition transcription and feature similarity from different text instances in adjacent frames, to join pairs with the same text instance.
Besides, these methods only deal with temporal information in the adjacent frames, they are not favorable to long temporal text tracking and recognition, because it is difficult to ensure the match correct of the same text instance in each adjacent frame.
Thus, a simple, effective end-to-end VTS framework with long-range modeling is highly desirable.

Here, we take a deeper look at the video text spotting tasks. The core of the task is that how to better obtain a steady tracking trajectory and recognition results for the same text instance in consecutive video frames. 
Recently, many transformer-based works~\cite{wang2021transformer,carion2020end,zhu2020deformable,zeng2021motr} have shown the potential of the Transformer structure~\cite{vaswani2017attention} for modeling long-range dependencies and learning temporal information across multiple frames.
Therefore, we attempt to explore a simple, but efficient video text spotting paradigm based on transformer.

In this paper, motivated by the query concept from MOTR~\cite{zeng2021motr}, we firstly propose an end-to-end trainable video text \textbf{D}Etection, \textbf{T}racking and \textbf{R}ecognition framework with Transformer~(\detr), which views the video text spotting task as a direct end-to-end long-range temporal variation problem, not the previous match problem.
We propose a new concept \textit{text query} and one temporal tracking loss to model video text spotting task over long-range sequence. 
Each text query is responsible to predict the entire track trajectory and the content of a text.
Since each text query follows the same text object all the time once the query is matched by one object, the paradigm can naturally remove the matching associations in adjacent frame and some hand-crafted components, \eg{NMS}.
During inference phrase, without hand-crafted match strategies, \detr use the classification scores of each text prediction to determine when the text appears and disappears.
As shown in Figure.~\ref{fig2}, \detr is a simple yet effective end-to-end framework, which includes three parts, \eg{} backbone, transformer encoder-decoder, and the recognition head with rotated RoI.
Encoder-decoder naturally models the tracking box with text queries, and the recognition head with rotated RoI enables the end-to-end training.
Given a video clip that consists of multiple frames as input, the \detr outputs the sequence of detection, tracking, and recognition results for each text instance in the video directly.
%

%
Our main contributions are summarized as follows:
\begin{itemize}
\itemsep -0cm
    \item 
    We firstly propose an end-to-end trainable video text spotting framework with Transformer, named \detr, 
    which simultaneously solve text detection, tracking, and recognition tasks \textit{seamlessly and naturally} in one unified framework. Without hand-crafted components(\eg{} \textit{the match of objects in adjacent frames, NMS}), the framework is significantly different from existing approaches, considerably simplifying the overall pipeline.
    
    \item 
    Different from the tracking-by-match paradigm over the adjacent frame, \detr views the VTS task as a \textit{long-range}~( more than 7 frames) temporal variation problem, which tracks and recognizes the same text in sequence with one text query.

    \item
    The proposed \detr achieves the state-of-the-art performance on four datasets for multi-stage tasks~(\ie{} text detection, tracking and spotting). Especially, \detr achieves $72.8\%$ ${\rm ID_{F1}}$ for video text spotting task on ICDAR2015~\cite{karatzas2015icdar}, with $11.3\%$ improvements than the previous SOTA methods.
    
\end{itemize}

\section{Related work}

\subsection{Text Spotting in Image}
Various methods~\cite{li2017towards,lyu2018mask,liu2020abcnet} based on deep learning have been proposed and have improved the performance considerably for image-level text spotting. 
Li \textit{et al.}~\cite{li2017towards} proposed the first end-to-end trainable scene text spotting method, which successfully uses an RoI Pooling~\cite{ren2015faster} to joint detection and recognition features. 
Its improved version~\cite{wang2021towards} further significantly improves the performance. Liao \textit{et al.}~\cite{lyu2018mask} propose a Mask TextSpotter which subtly refines Mask R-CNN and uses character-level supervision to detect and recognize characters simultaneously. 
Liu \textit{et al.}~\cite{liu2018fots} adopt an anchor-free mechanism to improve both the training and inference speed, and proposed an RoIRotate to enable the end-to-end text spotting training.
PAN++~\cite{wang2021pan++} proposed an end-to-end text spotting framework with masked roi, which reformulates a text line as a text kernel surrounded by peripheral pixels.
However, these image-based methods can not obtain temporal information in the video, which is essential for some downstream tasks such as video understanding.

\subsection{Text Spotting in Video}
Compared to text spotting algorithms in the image, there are rare video text spotting methods. 
Yin \textit{et al.}~\cite{yin2016text} provide a detailed survey, summarizes and compares the existing text detection, tracking, and recognition methods in a video before 2016. 
As shown in Fig.~\ref{fig2} (a), Wang \textit{et al.}~\cite{wang2017end} links texts in the current frame and several previous frames to obtain the tracking results by hand-crafted post-processing, such as IoU-based associations. 
As shown in Fig.~\ref{fig2} (b), Cheng \textit{et al.}~\cite{cheng2019you} propose a video text spotting framework by only recognizing the localized text one-time instead of frame-wisely recognition, which needs at least three-step with many hyper-parameters, including single-frame detection, aggregation, quality score.
As shown in Fig.~\ref{fig2} (c), Free~\cite{cheng2020free} is a improved version for faster and robuster video text spotting base on Cheng \textit{et al.}~\cite{cheng2019you}, which achieve better and faster performance.
Nguyen \textit{et al.}~\cite{nguyen2014video} performs character detection and recognition via scanning-window templates trained with mixture models.
Rong \textit{et al.}~\cite{rong2014scene} presents a hand-crafted framework of scene text recognition in multiple frames based on feature representation of scene text character~(STC) for character prediction and conditional random field~(CRF) model for word configuration.
As shown in Fig.~\ref{fig2} (d), Wu \textit{et al.}~\cite{wu2021bilingual} applies object features from the previous frame as a tracking query for the current frame, then uses IoU match to obtain tracking and recognition results.
The above methods include multiple models and hand-crafted components, which are complex pipelines and need multiple steps to generate the tracking trajectory and recognition results.
More detailed comparison are presented in Fig.~\ref{fig2}.
Besides, the above methods only can deal with temporal information in adjacent frames by text instance match, but it is difficult to ensure the match correct in each frame.

\subsection{Transformers in Vision}
Transformer was first proposed in~\cite{vaswani2017attention} as a new paradigm for the machine translation task. 
Since the 2020 year,
Transformers start to show promises in solving computer vision tasks. DETR~\cite{carion2020end,zhu2020deformable} propose a novel object detection paradigm based on Transformers, which largely simplifies the traditional detection pipeline, and achieves on excellent performances compared with CNN-based detectors~\cite{ren2015faster}.
Besides, there is a popularity of using transformer architecture in other vision tasks, such as segmentation~\cite{zheng2021rethinking}, object tracking~\cite{sun2020transtrack,zeng2021motr,zhao2022tracking} and even backbone construction~\cite{liu2021swin}.
Lately, some works~\cite{sun2020transtrack,zeng2021motr} show using a transformer in processing sequential visual data also make remarkable shots, especially for temporal tasks. 
MOTR~\cite{zeng2021motr} introduce a novel concept of track query and the contiguous query passing mechanism for multiple-object Tracking and long-range temporal relation. 
TransTrack~\cite{sun2020transtrack} track object by applying object features from the previous frame as a query of the current frame.
All the above methods prove the good capacity of transformer in modeling temporal relation, 
but for video text spotting or multi-orient text tracking, to the best of our knowledge, there are still no transformer-based solutions. 
Therefore, we propose a simple, but effective \detr method, which shows convincingly high performance on the popular benchmark.

\begin{figure*}[!t]
\begin{center}

\includegraphics[width=1\textwidth]{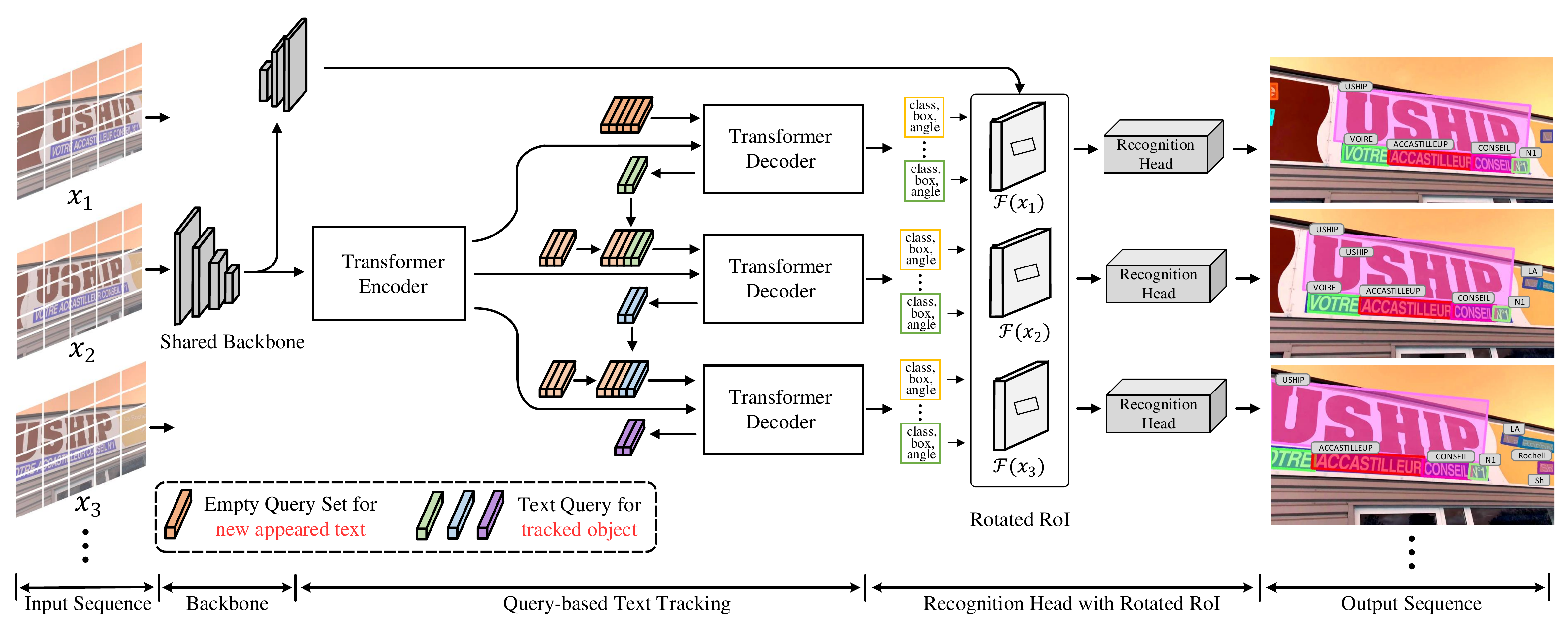}
\caption{\textbf{The overall architecture of \detr.} It contains three main components: 1). a backbone(\eg{}ResNet, PVT~\cite{wang2021pyramid}) is used to extract feature of video sequences; 2) a Transformer encoder models the relations of features, and a weight-shared decoder models each text move trajectory by text query. For the initial $F_{1}$ frame, an empty query set~(yellow box) is injected into the decoder to localize the initial objects and generate the initial text query for the next frame~($F_{2}$). For the $F_{t}$ frames, the specific text set from the previous frame is concatenated with an empty query set to generate the text query for the current frame; 3) an recognition head with Rotated RoI is designed to recognize text.}
\label{architecture}
\end{center}
\vspace{-0.5cm}
\end{figure*}

\section{Our Method: \detr}
The proposed method tackles the video text spotting task by modeling it as a direct sequence prediction problem, where each text query is responsible to predict an entire text instance trajectory and the corresponding text content. 
To achieve this goal, two ingredients are essential: 
(1) \textit{The simple pipeline}: without multiple models and hand-crafted strategies(\eg{} NMS, Track Post-Processing), \detr tracks and recognizes text instances during continuous video frames directly.
The whole pipeline includes three parts: backbone, transformer-based encoder and decoder, and recognition head with Rotated RoI.
(2) \textit{Temporal Tracking Loss for Text Query}: for each frame, there are two kinds of text tracking losses, \ie{} tracked object with the same query(text query) from the previous frames and new appeared text object with empty query. 
Temporal Tracking Loss is a sequence-aware tracking loss for modeling long-range temporal trajectory.

\subsection{\detr architecture}
The overall \detr architecture is depicted in Figure~\ref{architecture}, which contains three main components: A backbone~(\eg{} ResNet, PVT~\cite{wang2021pyramid}) is used to extract feature representation, 
transformer-based encoder network and decoder network models the relations of pixel-level features and detects new-born text instance and trace already existed instance,
the attention-based recognition head with Rotatedt RoI is utilized to obtain the final recognition results.

\textbf{Backbone}. The backbone extracts the original pixel-level feature sequence of the input video clip. 
Assume that the initial video clip with $n$ consecutive frames is denoted by $x_{\rm t}\in{\{x_1,x_2,...,x_n\}}$.
A transformer
backbone~\cite{wang2021pyramid} is used to generate the corresponding lower-resolution activation map $f_{\rm t} \in{\{f_1,f_2,...,f_n\}}$.

\textbf{Transformer Encoder.} We adopt deformable transformer encoder~\cite{zhu2020deformable} to model the similarities among all the pixel level features for the extracted features $f_{\rm t}$. The corresponding feature output for each frame can be denoted by $D(f_{\rm t})$, where $t$ corresponds to $t$-th frame. 

\textbf{Transformer Decoder.} 
The decoder follows the standard architecture of the deformable DETR~\cite{zhu2020deformable}, which decodes the top pixel features that can represent the text instances of each frame.
The difference with the deformable DETR decoder is that our $N$ queries are composed of two parts: empty queries $Q_{\noobject}$ and text queries $Q_{text}$ from the previous frame, as shown in Figure~\ref{architecture}.
For the first frame $x_1$, the corresponding feature $D(f_{\rm 1})$ together with the fixed-length empty queries $Q(\noobject)$ are input to the transformer decoder to generate the initialized text object and text queries $Q_{1}$. 
And then, we concatenate text queries $Q_{1}$ and empty queries $Q(\noobject)$ to generate a new query set for the $x_{2}$ frame.
Similarly, for any $x_{t}$~($t>1$) frame, the corresponding $N$ text instance queries are composed of text queries $Q_{t-1}$ from the previous frame and empty queries $Q(\noobject)$. 

\textbf{Rotated RoI Align.} To enable end-to-end training, similar to RRPN~\cite{ma2018arbitrary}, RRPN++~\cite{ma2020rrpn++}, we adopt the Rotated Region-of-Interest~(Rotated RoI) to extract the features of each text from the output feature map of upsampling. 
%
%
The whole process can be divided into three steps.
1) We firstly upsample and concatenate low-level feature maps and high-level semantic feature maps $f_{\rm t} \in{\{f_1,f_2,...,f_4\}}$ from the shared backbone to obtain the feature maps~$F$, whose size is same as that of the input image. 
2) Translation matrix~$\textbf{T}$ and rotation matrix~$\textbf{R}$ parameters for affine transformation are computed via predicted or ground truth coordinates of text proposals.
3) Affine transformations are applied to the feature map~$F$ for each region respectively, and horizontal feature maps of text regions are obtained.
For the last two step, given one feature map and the text region-of-interest representation~($c_x, c_y, c_h,c_w, \theta $). $c_x, c_y, c_h,c_w, \theta $ denotes the center coordinate, height, width, and orientation angle of the Rotated RoI, respectively.
It is easy to produce the final Rotated RoI feature using the affine transformation:

\begin{footnotesize}
\begin{equation}
\begin{aligned}
    \left[
\begin{array}{c}
x'\\y'\\1
\end{array} \right]
&=\mathbf{R}\mathbf{T}
    \left[
\begin{array}{c}
x\\y\\1 
\end{array} \right] \\
&= \left[
\begin{array}{ccc}
\cos\alpha & \sin\alpha & 0\\
-\sin\alpha & \cos\alpha & 0\\
0 & 0 & 1
\end{array} \right] 
\left[
\begin{array}{lll}
1 & 0 & -c_x\\
0 & 1 & -c_y\\
0 & 0 & 1
\end{array} \right]
\left[
\begin{array}{c}
x\\y\\1
\end{array} \right]
\end{aligned}
\end{equation}
\end{footnotesize}
where $(x,y)$ is a point in the input feature map, and $(x',y')$ refers to the corresponding points in the rotated feature map.
The final horizontal feature of each text is extracted from the final Rotated RoI feature, and resized to the fixed size~($h\times w$).

\textbf{Recognition Head.} Benefiting from the shared backbone feature and Rotated RoI, we design an attention-based recognition head for end-to-end training.
Following the PAN++~\cite{wang2021pan++}, our head is also a seq2seq model composed of a starter and a decoder.
The starter produces SOS’s feature vector with a linear transformation (embedding layer), which is the input of the decoder at the initial time step.
The decoder predicts symbol $y_{rec}$ with two LSTM layers and one multi-head attention layer at each time step, and the output symbol of the previous step is fed into later LSTM until predicting the end of the string (EOS) token.
The attention-based recognition head is a basic and simple recognition network, which enables the model to learn a character-level language model representing output class dependencies.

\subsection{Temporal Tacking Loss for Text Query}

\textbf{Multi-orient Box Matching for Single Frame.}
To fit the rotated text instances, similar to TransVTSpotter~\cite{wu2021bilingual}, we use the multi-orient box bipartite matching for detecting and tracking rotated scene text, where the core idea is an angle prediction of a rotated bounding box. 
Similar to DETR~\cite{carion2020end}, the proposed method infers a fixed-size set of $N$ predictions, then produces an optimal bipartite matching between predicted and ground-truth objects, finally optimizes object-specific (rotated bounding box) losses.
For $t$-th frame, let us denote the ground truth set of objects by $y^t$, and $\hy = \{\hy_i^t\}_{i=1}^{N}$ the set of $N$ predictions. 
To find a bipartite matching between these two sets, we search for a permutation of $N$ elements $\sigma \in \Sigma_N$ with the lowest cost:
\begin{equation}
\label{eq:matching}
    \hat{\sigma}^t = \argmin_{\sigma\in\Sigma_N} \sum_{i}^{N} \lmatch{y_i^t, \hy^t_{\sigma(i)}},
\end{equation}
where $\lmatch{y_i^t, \hy^t_{\sigma(i)}}$ is a pair-wise \emph{matching cost}~(Hungarian algorithm~\cite{kuhn1955the}) between ground-truth $y_i$ and a prediction with index $\sigma(i)$ in $t$-th frame. 
%
%
Each element $i$ of the ground-truth set can be seen as a $y_i = (c_i, b_i, a_i)$ where $c_i$ is the target class label, $b_i \in [0, 1]^4$ is a vector that defines ground-truth box center coordinates and its height and width relative to the image size, and $a_i$ is rotation angle between the longest edge of arbitrary-oriented box and horizontal line~(x-axis). 
For the prediction with index $\sigma(i)$ we define probability of class $c_i$ as $\hp_{\sigma(i)}(c_i)$, the predicted box as $\hb_{\sigma(i)}$, and the predicted angle as $\ha_{\sigma(i)}$. 
Thus we define
$\lmatch{y_i^t, \hy^t_{\sigma(i)}}$ as:
\begin{equation}
\begin{aligned}
\lmatch{y_i^t, \hy^t_{\sigma(i)}} &= -\indic{c_i\neq\noobject}\hp^t_{\sigma(i)}(c_i) \\
&+ \indic{c_i\neq\noobject} \bloss{b^t_{i}, \hb^t_{\sigma(i)}} \\
& +\indic{c_i\neq\noobject} \aloss{a^t_{i}, \ha^t_{\sigma(i)}}\,. 
\end{aligned}
\end{equation}

Finally, we could compute the detection loss function with all pairs matched at the $t$-th frame:

\begin{equation}
\begin{aligned}
\hloss{y^t, \hy^t} &= \sum_{i=1}^N \Bigl[-\log  \hp^t_{\hat{\sigma}(i)}(c_{i}) \\
&+ \indic{c_i\neq\noobject} \bloss{b^t_{i}, \hb^t_{\hat{\sigma}}(i)} \\
& +\indic{c_i\neq\noobject} \aloss{a^t_{i}, \ha^t_{\sigma(i)}}\Bigl]\,, 
\label{equation3}
\end{aligned}
\end{equation}
where $\bloss{\cdot}$ is a linear combination of the $\ell_1$ loss and the generalized IoU loss~\cite{rezatofighi2019generalized,carion2020end}. Specifically, the linear combination is defined by $\lambda_{\rm iou}\iouloss{b_{i}, \hb_{\sigma(i)}} + \lambda_{\rm L1}||b_{i}- \hb_{\sigma(i)}||_1$ where $\lambda_{\rm iou}, \lambda_{\rm L1}\in\Re$ are hyperparameters. And $\aloss{\cdot}$ refers to a cosine angle embedding loss $1-cos(\ha_{\sigma(i)}-a_{i})$, enabling the network to fit rotated text instances. 

\textbf{Temporal Tracking Loss over Multiple Frames.}
Multi-oriented box matching solve rotated scene text detection by the rotation angle prediction, how to track each text instance with the same text query accurately becomes another vital problem.
Motivated by DETR~\cite{carion2020end,zhu2020deformable} and MOTR~\cite{zeng2021motr}, we propose a concept \textit{text query} to model video text spotting over long-range sequence. 
If the match in Equation.~\ref{eq:matching} is successful, the matched text query $Q_{text}$ is responsible to track the corresponding text instance until it disappears.  

To achieve this goal, the fixed-size set of $N$ queries needs to contain two parts: empty queries $Q(\noobject)$ for detecting the new-born text instances and text queries $Q_{text}$ for tracking already tracked text instances.
\textbf{For new-born text instances}, an optimal bipartite matching in Equation.~\ref{eq:matching} is produced between prediction from empty queries $Q(\noobject)$ and ground truth  of new-born text instances. 
\textbf{For already tracked text instances}, without the optimal bipartite matching, the already matched text queries $Q_{text}$ are inherited from the previous frame.
Finally, the matched empty queries $Q(\noobject)$ and the inherited text queries $Q_{text}$ are concatenated together and input to the Transformer decoder to update the representation in the current frame.
For $t$-th frame, the fixed-size of $N$ queries can be formulated as:

\begin{equation}
\begin{split}
    Q^t_{text}=\left\{\begin{array}{lcl}
      Q^t_{\sigma(i)}(\noobject)  &  & t = 1
      \\
      Q^t_{\sigma(i)}(\noobject)\cup Q^{t-1}_{text}   &  & t\geq 2
    \end{array}\right. 
\end{split} 
\label{label_assign_tracked}
\end{equation}
where $\sigma(i)$ refers to the lowest matching cost index from \textit{Hungarian algorithm} in Equation.~\ref{eq:matching}. For the initial frame($t=1$), since no already matched text queries $Q_{text}$ from the previous frame, empty query set with index $\sigma(i)$ is used to localize the initial objects and generate the initial text query set for the next frame($t=2$). 
For $t$-th frame($t>=2$), the query set contain two parts. Empty query set with index $\sigma(i)$ from the pair-wise matching is used to detect new-born text instances. Text query set $Q_{text}$, as the already matched in the previous frame, track already tracked text instances.

Different from the previous methods~\cite{cheng2020free,wang2017end} training with single frame without temporal information, \detr takes a video clip~(around 8 frames) as input to learn long-range temporal variances. The overall loss in the whole video clip sequence($T$ frames in total) could be computed:

\begin{equation}
\begin{split}
    \hspace{-1.6mm}\mathcal{L}_{track}(y, \hy) \hspace{-0.5mm}=\hspace{-0.5mm} \frac{\sum\limits_{t=1}^{T}\hspace{-0.5mm}(\hloss{y^t, \hy^t})\hspace{-0.8mm}+\hspace{-0.8mm}\mathcal{L}(y^t, \hy^t_{Q^{t-1}_{text}})\hspace{-0.5mm}) }{\sum\limits_{t=1}^{T}(G^{t})}\,, 
\end{split}
\label{Eq8}
\end{equation}
where $G^{t}$ denotes the total number of the ground-truths at $T_{i}$ frame.
Similar to the $\hloss{y^t, \hy^t}$ in Equation.~\ref{equation3}, the $\mathcal{L}(y^t, \hy^t_{Q^{t-1}_{text}})$ is the detection loss of a single frame. The difference is it corresponding to the already matched text queries $Q^{t-1}_{text}$ and $\hat{\sigma}^{t-1}$ from the previous frame. 

\subsection{Loss Function}
The proposed pipeline mainly contains two losses, \ie{} temporal tracking loss and recognition loss, which are responsible for three different tasks, \ie{} text detection, tracking, and recognition tasks.
The whole loss function can be formulated as Equation.~\ref{eq:loss1}:

\begin{equation}
\label{eq:loss1}
    \mathcal{L}=\sigma _1 \mathcal{L}_{track} + \sigma _2 \mathcal{L}_{rec} ~,
\end{equation}
where $\sigma _1$ and $\sigma _2$ are two weight parameters. 
The loss function $\mathcal{L}_{rec}$ of text recognition refers to Cross-Entropy loss between ground truth and predicted words. $\mathcal{L}_{track}$ is the temporal tracking loss composed of two parts, and each part includes three su-losses, \eg{} class loss, rotation angle loss $\aloss{\cdot}$, and generalized IoU loss $\bloss{\cdot}$, as shown in Equ.~\ref{Eq8}.

\subsection{Inference}
During inference, many hand-crafted components and hyper-parameter, \eg{} explicit track associations, NMS are no longer required.
The fixed-size set of $N$ queries are used to predict a set of text moving trajectories and the corresponding content.
And we use classification scores of prediction $\ha_{\sigma(i)}$ to determine when a text instance appears and disappears.
A text instance appears while classification scores are higher than one threshold $\tau$.
In turn, we remove the text instance while its classification score is lower than the threshold $\tau$.
In all experiments, we set the threshold $\tau$ to $0.5$.

\begin{table*}[htb!]
    \centering
    \small 
    \setlength{\tabcolsep}{1mm}
    \caption{\textbf{Multi-Oriented Box Matching(MOBM) for detection and tracking task on ICDAR2015video~\cite{karatzas2015icdar}}. Cosine loss brings more gains than L1 loss.}
    \footnotesize
\scalebox{1.0}{
\begin{tabular}{l | ccc | ccc}
    Method & Precision  & Recall & F-measure & ${\rm ID_{F1}}$$\uparrow$  & ${\rm MOTA}$$\uparrow$ & ${\rm MOTP}$$\uparrow$\\
    \hline
    ours,w/o MOBM & 67.7 & 56.1 & 61.5 & 41.1 & 18.4 & 68.7 \\
    ours,w/ MOBM($L1$) & 77.4 & \textbf{70.5} & 73.8 & 63.3 & 45.6 & \textbf{76.1}  \\
    ours,w/ MOBM($Cosine$) & \textbf{80.6} & 70.2 & \textbf{75.0} & \textbf{65.5} & \textbf{47.7} & 74.1 \\
    
\end{tabular}}
    \label{ablation}
    
\end{table*}

\begin{table*}[t]
    \centering
    \small 
    \setlength{\tabcolsep}{1mm}
    \caption{\textbf{Effect of Temporal Tracking Loss(Text Query-based) for Text Tracking Task.} Deformable-DETR~\cite{zhu2020deformable} is used as the base detector for text tracking on ICDAR2015 (video)~\cite{karatzas2015icdar}. The related settings of IoU and Transcription-based follows the work~\cite{wang2017end}.}
    \footnotesize
\begin{tabular}{l|ccc}
Method & ${\rm ID_{F1}}$$\uparrow$  & ${\rm MOTA}$$\uparrow$ & ${\rm MOTP}$$\uparrow$ \\
\shline
\hline
IoU-based & 56.3 & 43.1 & 75.8 \\
Transcription-based & 56.9 & 43.5 & \textbf{75.9}  \\
Temporal Tracking Loss for Text Query & \textbf{65.5} & \textbf{47.7} & 74.1  \\

\end{tabular}

    \label{tab:query}
\end{table*}
\begin{table}[t]
    \centering
    \small 
    \setlength{\tabcolsep}{1mm}
    \caption{\textbf{The input sequence length for text tracking task on ICDAR2015~(video)~\cite{karatzas2015icdar}.} Long-range temporal information brings huge gains.}
    \footnotesize
\begin{tabular}{l|ccccc}
    Length & ${\rm ID_{F1}}$$\uparrow$  & ${\rm MOTA}$$\uparrow$ & ${\rm MOTP}$$\uparrow$ & ${\rm M\mbox{-}Tracked}$$\uparrow$ & ${\rm M\mbox{-}Lost}$$\downarrow$ \\
    \shline
    \hline
    3 & 61.2 & 44.8 & 74.1 & 38.9 & 34.5 \\
    4 & 63.4 & 45.5 & 74.1 & 40.1 & 33.2 \\
    5 & 65.2 & 47.3 & 74.0 & 42.0 & 32.5 \\
    6 & \textbf{65.5} & \textbf{47.7} & \textbf{74.1} & \textbf{42.0} & \textbf{32.1} \\
    
    \end{tabular}
    
    \label{ablation2}
\end{table}

\begin{figure*}
\centering
\begin{minipage}[h]{0.48\textwidth}
\centering
\includegraphics[width=9cm]{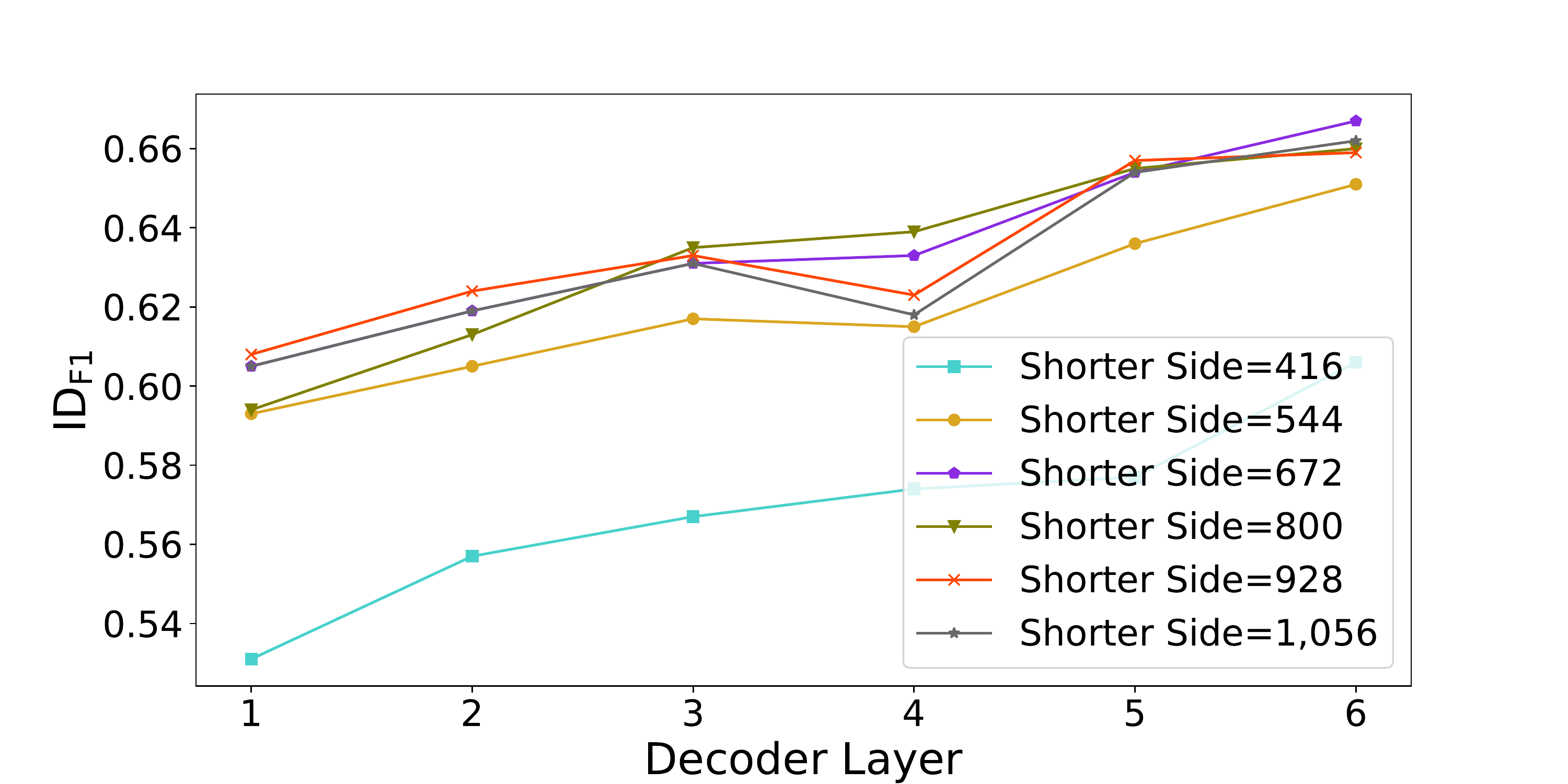}
\subcaption{\textbf{Effect of decoder layer number and the shorter side of input image.} `Shorter Side' denotes the length of image shorter side.}
\label{Visualization1}
\end{minipage}
\begin{minipage}[h]{0.48\textwidth}
\centering
\includegraphics[width=9cm]{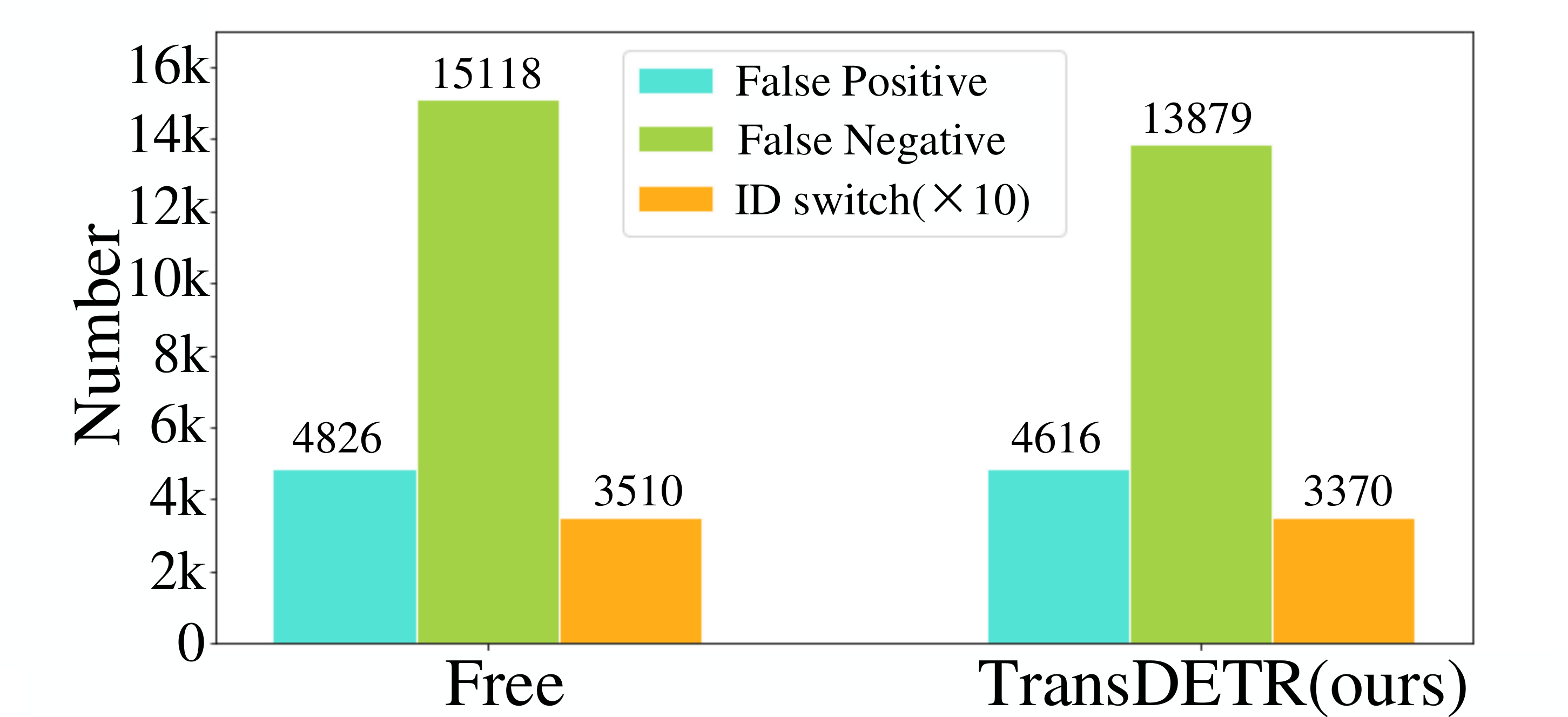}
\subcaption{\textbf{Comparison with Free~\cite{cheng2020free} in FPs, FNs and ID switch for text tracking task on ICDAR2013~\cite{karatzas2013icdar}.} `$\times10$' denotes $10$ times.}
\label{Visualization3}
\end{minipage}
\end{figure*}

\section{Experiments}

\subsection{Datasets}
\label{dataset}
\textbf{ICDAR2013 Video~\cite{karatzas2013icdar}} is proposed in the ICDAR2013 Robust Reading Competition, which contains 13 videos for
training and 15 videos for testing. These videos are harvested from indoors and outdoors scenarios, and each text is labeled as a quadrangle with 4 vertexes in word-level.

\textbf{ICDAR2015 Video~\cite{karatzas2015icdar}} is the expanded version of ICDAR2013~(video), which consists of a training set of 25 videos (13,450 frames) and a test set of 24 video (14,374 frames). Similar to ICDAR2013~(video), text instances in this dataset are labeled at the word level. Quadrilateral bounding boxes and transcriptions are provided. 

\textbf{Minetto~\cite{minetto2011snoopertrack}} contains 5 videos in outdoor scenes. The resolution is fixed as 640×480. Each text is labeled in the form of axis-aligned bounding box. As all videos are for testing, we use the model trained on ICDAR2015 Video to evaluate this dataset directly.

\textbf{YouTube Video Text~(YVT)~\cite{nguyen2014video}} dataset is harvested from YouTube, contains $30$ videos, where $15$ videos for training and $15$ videos for testing. Different from the above datasets, it contains web videos except for scene videos.

\subsection{Implementation Details}
All the experiments are conducted on PyTorch with Tesla V100 GPUs. We use the MOTR~\cite{zeng2021motr} as our basic network. 
For all experiments in our paper, the model is firstly pretrained on the COCO-Text~\cite{veit2016coco}, and then finetune on other video datasets. 
COCO-Text is the largest scene text detection dataset with 63,686 images, which reuses the images from MS-COCO dataset~\cite{lin2014microsoft}.
For static images from COCO-Text, we apply the random shift~\cite{zhou2020tracking} to generate video clips with pseudo tracks.
AdamW~\cite{loshchilov2017decoupled} as the optimizer for total 20 epochs with the initial learning rate of 2e\mbox{-}4. The learning rate decays to 2e\mbox{-}5 at 10 epochs.
The batch size is set to be 1, where each batch contains 8 frames. The data augmentation includes random horizontal, scale augmentation, resizing the images whose shorter side is by 480-800 pixels while the longer side is by at most 1333 pixels.
Following the setting in MOTR~\cite{zeng2021motr}, we randomly sample key frames from video sequences with an interval to solve the problem of different frame rates, where the max interval is set to 5 in our experiments.
All inference speed performances are tested with a batch size of 1 on a V100 GPU and a 2.20GHz CPU in a single thread.
In the metric of the paper, Mostly Tracked~($M\mbox{-}Tracked$) refers to the number of text instance tracked for at least 80 percent of lifespan, Mostly Lost~($M\mbox{-}Lost$) denotes the number of objects tracked less than 20 percent of lifespan in tracking.

\textbf{Inconsistency from Evaluation Metric.} 
The inconsistency between the reported performance and the original paper lies in the update of the evaluation metric.
From 2020, Robust Reading Competition Website~\footnote{https://rrc.cvc.uab.es/?ch=3\&com=evaluation\&task=1} has updated the evaluation method, and the old metric and ground truth is not available. 
All works have already used the new metric after 2020.
As for ICDAR2015 Video, we test it via the online evaluation on Robust Reading Competition Website.
Other datasets are evaluated by the offline metric, and we have provided the metric in our open source code.

\subsection{Ablation Study}
Here, we conduct five groups of ablation experiments to study the core factors of \detr. 

\textbf{Multi-Oriented Box Matching.} To evaluate the importance of the multi-oriented box matching cost, we train several models turning it $\aloss{a^t_{i}, \ha^t_{\sigma(i)}}$ on and off. 
As shown in Table~\ref{ablation}, for detection and tracking tasks on ICDAR2015~\cite{karatzas2015icdar}, we train a model without multi-oriented box matching and angle prediction, a model with $L1\mbox{-}based$ angle loss, and a model with $Cosine\mbox{-}based$ angle loss.
With multi-oriented box matching~(cosine loss), the model achieves $13.5\%$ improvements of F-measure than the counterpart without the matching for the text detection task, $24.4\%$ ${\rm ID_{F1}}$ improvements for text tracking task on ICDAR2015 video.

\textbf{Temporal Tracking Loss.}
Query-based temporal text loss plays a vital role in the proposed video text tracker. As shown in Table~\ref{tab:query}, we evaluate the effectiveness by comparing with other association tracking methods, \eg{} IoU-based association tracker and transcription-based association tracker.
Following the work~\cite{wang2017end}, IoU-based and transcription-based association trackers link and match text objects in adjacent frames by IoU and edit distance of text instances. 
The query-based method with temporal text loss shows a significant improvement than the other methods, especially for ${\rm ID_{F1}}$, which achieves up to $8.6\%$ improvement. 
The previous tracing-by-match paradigm requires successful text match in each adjacent frame, which is difficult to achieve.
Matching failure in any adjacent frames would lead to the ID switching. 
The query-based model temporal text instances relationship during multi-frame, instead of matching text instance in the adjacent frames.
The novel paradigm enhances the tracking stably by alleviating the problem of ID switching and object missing.

\textbf{Video Sequence Length.} The input video sequences of different lengths make a different impact on temporal relation modeling. 
As shown in Table~\ref{ablation2} (c), we experiment with various video sequence lengths to explore the effect of the sequence length setting.
The longer input video sequences, the better performance, the video sequence with six frames shows $4.3\%$ improvement than the video of three frames.
With the increasing of video sequence length, the model presents a better performance, and achieves the best with $65.5\%$ ${\rm ID_{F1}}$ and $47.7\%$ ${\rm MOTA}$.
Therefore, we argue that the longer video sequence with more frames contributes to better temporal relation modeling.

\textbf{Decoder Layer and Shorter Side of Input Image.} We evaluate the importance of decoder layer number and the image shorter side in Figure~\ref{Visualization1}. 
With the increasing of decoder layer number, the performance presents a gradual improvement, achieving the highest accuracy with $6$ layers.
Similarly, better performance is presented with a shorter side of image increasing. 
And the performance drops dramatically at least $5\%$ while the shorter side is less than $500$ pixels, where much low-level feature information lost. 
In all experiments, we set the decoder layer number and shorter side of the image to $6$ and $800$ for the trade-off between accuracy and speech, respectively.

\textbf{Efficiency for fixing ID switching, FP and FN.}
To further present the efficiency of \detr for text tracking task, we summarize and compare the number of false-positive, false-negative, and ID switches on ICDAR2013~\cite{karatzas2013icdar} in Figure~\ref{Visualization3}.
Compared with Free~\cite{cheng2020free}, our \detr suppresses obviously false positive, false negative, and ID switch by long-sequence modeling.
Especially for false negatives, the improved performance is quite remarkable, with a 1,239 false negatives decrease.
We argue that the improvements mainly come from these undetected texts in the middle frames.
The previous paradigms rely on the image detection model, which is difficult to obtain a steady detection performance for all text instances during long-sequence video frames. 
Our \detr searches these texts with low confidence by query-based long-sequence modeling, which off huge dependence on detection model.

\begin{table}[t]
    \centering
    \small 
    \setlength{\tabcolsep}{2mm}
    \caption{\textbf{Video text detection performance on ICDAR2013(video)~\cite{karatzas2013icdar}.} \detr presents better performance and faster inference speed.}
\footnotesize
\begin{tabular}{l|ccc|c}
\multirow{2}{*}{Method} & \multicolumn{3}{c|}{Video Text Detection/\%} & \multirow{2}{*}{FPS}\\
\cline{2-4} 
& Precision  & Recall & F-measure & \\
\shline
\hline
Epshtein \textit{et al.}~\cite{epshtein2010detecting} & 39.8 & 32.5 & 35.9 & -\\
Zhao \textit{et al.}~\cite{zhao2010text} & 47.0 & 46.3 & 46.7 & -\\
Yin \textit{et al.}~\cite{yin2013robust} & 48.6 & 54.7 & 51.6 & -\\
Khare \textit{et al.}~\cite{khare2017arbitrarily} & 57.9 & 55.9 & 51.7 & -\\
Wang \textit{et al.}~\cite{wang2018scene} & 58.3 & 51.7 & 54.5 & -\\
Shivakumara \textit{et al.}~\cite{shivakumara2017fractals} & 61.0 & 57.0 & 59.0 & -\\
Wu \textit{et al.}~\cite{wu2015new} & 63.0 & 68.0 &  65.0 & -\\
Yu \textit{et al.}~\cite{yu2021end} & \textbf{82.4} & 56.4 & 66.9 & -\\
Wei \textit{et al.}~\cite{feng2021semantic} & 75.5 & 64.1 & 69.3 & -\\
Free~\cite{cheng2020free} & 79.7 & 68.4 & 73.6 & 8.8\\

\detr(ours) & 80.6 & \textbf{70.2} & \textbf{75.0} & \textbf{15.4} \\

\end{tabular}

    \label{tab:ICDAR2013det}
\end{table}

\begin{figure*}[t]
\begin{center}

\includegraphics[width=1\textwidth]{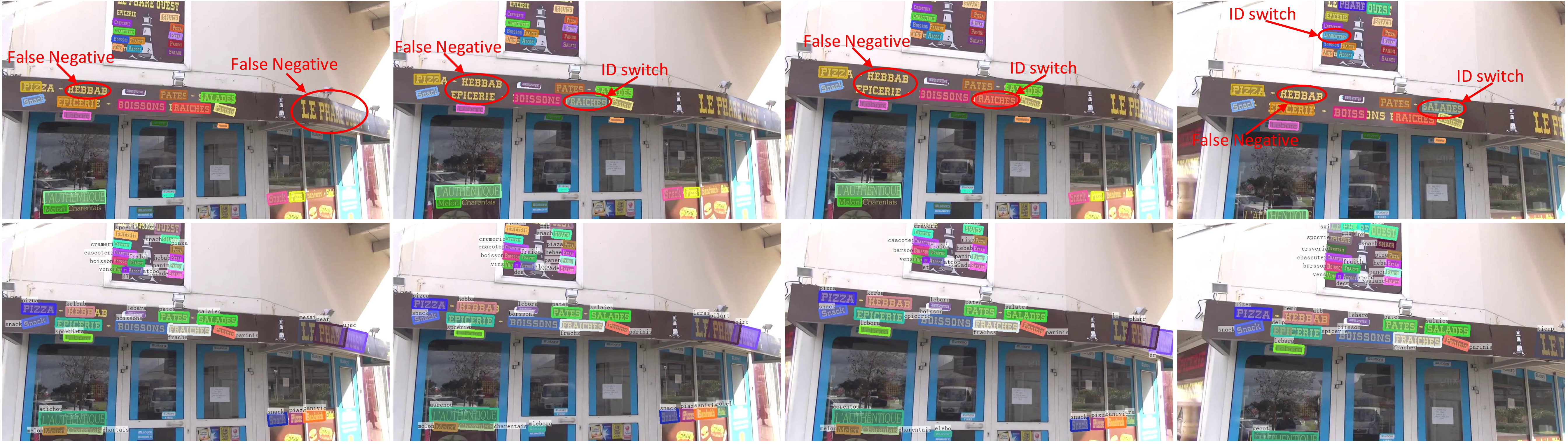}
\caption{\textbf{Tracking and spotting Visualization for different methods.} Compared with Free~\cite{cheng2020free}~(top row), our \detr(bottom row) shows better performance for solving false positive and ID switch problems.}
\label{Visualization}
\end{center}
\end{figure*}

\begin{table*}[t]
    \centering
    \setlength{\tabcolsep}{1mm}
    \caption{\textbf{Text tracking performance on YVT(video)~\cite{nguyen2014video}, ICDAR2013(video)~\cite{karatzas2013icdar} and Minetto~\cite{minetto2011snoopertrack}}. `M-Tracked' and `M-Lost' denote `Mostly Tracked' and `Mostly Lost', respectively. $^*$ denotes our implement with the official code and weight.}
\footnotesize 
\begin{tabular}{l|c|ccccc|c}
\setlength{\tabcolsep}{0.6mm}
\multirow{2}{*}{Dataset} & \multirow{2}{*}{Method} & \multicolumn{5}{c|}{Video Text Tracking/\%} & \multirow{2}{*}{FPS}\\
\cline{3-7} 
& & ${\rm ID_{F1}}$$\uparrow$  & ${\rm MOTA}$$\uparrow$ & ${\rm MOTP}$$\uparrow$ & ${\rm M\mbox{-}Tracked}$$\uparrow$ & ${\rm M\mbox{-}Lost}$$\downarrow$ &  \\
\shline \hline

\multirow{7}{*}{ICDAR2015\cite{karatzas2015icdar}} & 
USTB\_TexVideo~\cite{karatzas2015icdar} & 25.9 & 7.4 & 70.8 & 7.4 & 66.1 & -\\
~ & StradVision\mbox{-}1~\cite{karatzas2015icdar} & 25.9 & 7.9 &  70.2 & 6.5& 70.8 & -\\
~ & USTB(II\mbox{-}2)~\cite{karatzas2015icdar} & 21.9 & 12.3 &  71.8 & 4.8& 72.3 & -\\
~ & AJOU~\cite{koo2013scene} & 36.1 & 16.4 &  72.7 & 14.1& 62.0 & -\\
~ & Free~\cite{cheng2020free} & 57.9 & 43.2 &  \textbf{76.7} & 36.6 & 44.4 & 8.8\\
~ & TransVTSpotter~\cite{wu2021bilingual} & 57.3 & 44.1 & 75.8 & 34.3 & 33.7 & 9.0\\
~ & \cellcolor{mygray} \detr & \cellcolor{mygray} \textbf{65.5} & \cellcolor{mygray} \textbf{47.7} & \cellcolor{mygray} 74.1 &  \cellcolor{mygray} \textbf{42.0} & \cellcolor{mygray} \textbf{32.1} & \cellcolor{mygray} \textbf{16.7}\\
\hline

\multirow{6}{*}{Minetto\cite{minetto2011snoopertrack}} & Zuo \textit{et al.}~\cite{zuo2015multi} & - & 56.4 & 73.1 & - & - & -\\
~ & Pei \textit{et al.}~\cite{pei2018scene} & - & 73.1 & 57.7 & - & - & - \\
~ & AGD\&AGD\cite{yu2021end} & - & 75.6 & 74.7 & - & - & -\\
~ & Yu \textit{et al.}\cite{yu2021end} & - & 81.3 & \textbf{75.7}  & - & - & - \\
~ & TransVTSpotter\cite{wu2021bilingual} & 74.7 & 84.1 &  77.6 & - & - & 9.0\\
~ &\cellcolor{mygray}\detr & \cellcolor{mygray} \textbf{76.6} & \cellcolor{mygray} \textbf{86.5} & \cellcolor{mygray} 75.5 & \cellcolor{mygray} \textbf{94.3} & \cellcolor{mygray} \textbf{0} & \cellcolor{mygray} \textbf{18.2} \\
\hline

\multirow{2}{*}{ICDAR2013\cite{karatzas2013icdar}} & YORO$^*$~\cite{cheng2019you} & 62.5 & 47.3 & 73.7&  33.1& 45.3 & 14.3\\
~ & \cellcolor{mygray}\detr & \cellcolor{mygray}\textbf{67.2} & \cellcolor{mygray} \textbf{54.7} & \cellcolor{mygray} \textbf{76.6}  & \cellcolor{mygray} \textbf{43.5} & \cellcolor{mygray} \textbf{33.2} & \textbf{15.4} \cellcolor{mygray}\\
\hline

\multirow{2}{*}{YVT\cite{nguyen2014video}} & Free\cite{cheng2020free} & - & 54.0 & 78.0 & -& -&-\\
~ & \cellcolor{mygray}\detr & \cellcolor{mygray}\textbf{73.4} & \cellcolor{mygray}\textbf{55.7} & \cellcolor{mygray}76.2 & \cellcolor{mygray} \textbf{57.2} & \cellcolor{mygray} \textbf{33.8}& \cellcolor{mygray} \textbf{17.5}\\

\end{tabular}
    
    \label{tab:icdar13}
\end{table*}

\subsection{Comparison with State-of-the-arts}
We compare \detr against state-of-the-art methods in terms of three tasks, \ie{} video text detection, video text tracking and video text spotting. 

\textbf{Video text detection.}
Following previous works~\cite{cheng2020free}, ICDAR2013 video as a popular public dataset is used for use to verify the effectiveness of \detr.
Table~\ref{tab:ICDAR2013det} presents the video text detection performance on ICDAR2013~(video)~\cite{karatzas2013icdar}. 
\detr achieves \textit{the best result} among methods without any bells and whistles.
Compare with Free~\cite{cheng2020free} using a powerful base detector EAST~\cite{zhou2017east}, our \detr still achieves $1.4\%$ F-measure improvement without hand-crafted components such as NMS.
Therefore, we consider that the query-based temporal modeling from our \detr can help the network to learn a better feature representation for video text detection.
Note: we do not evaluate the performance on ICDAR2015 video due to the test set annotation of that is not available.

\begin{figure*}[t]
\begin{center}
\includegraphics[width=1\textwidth]{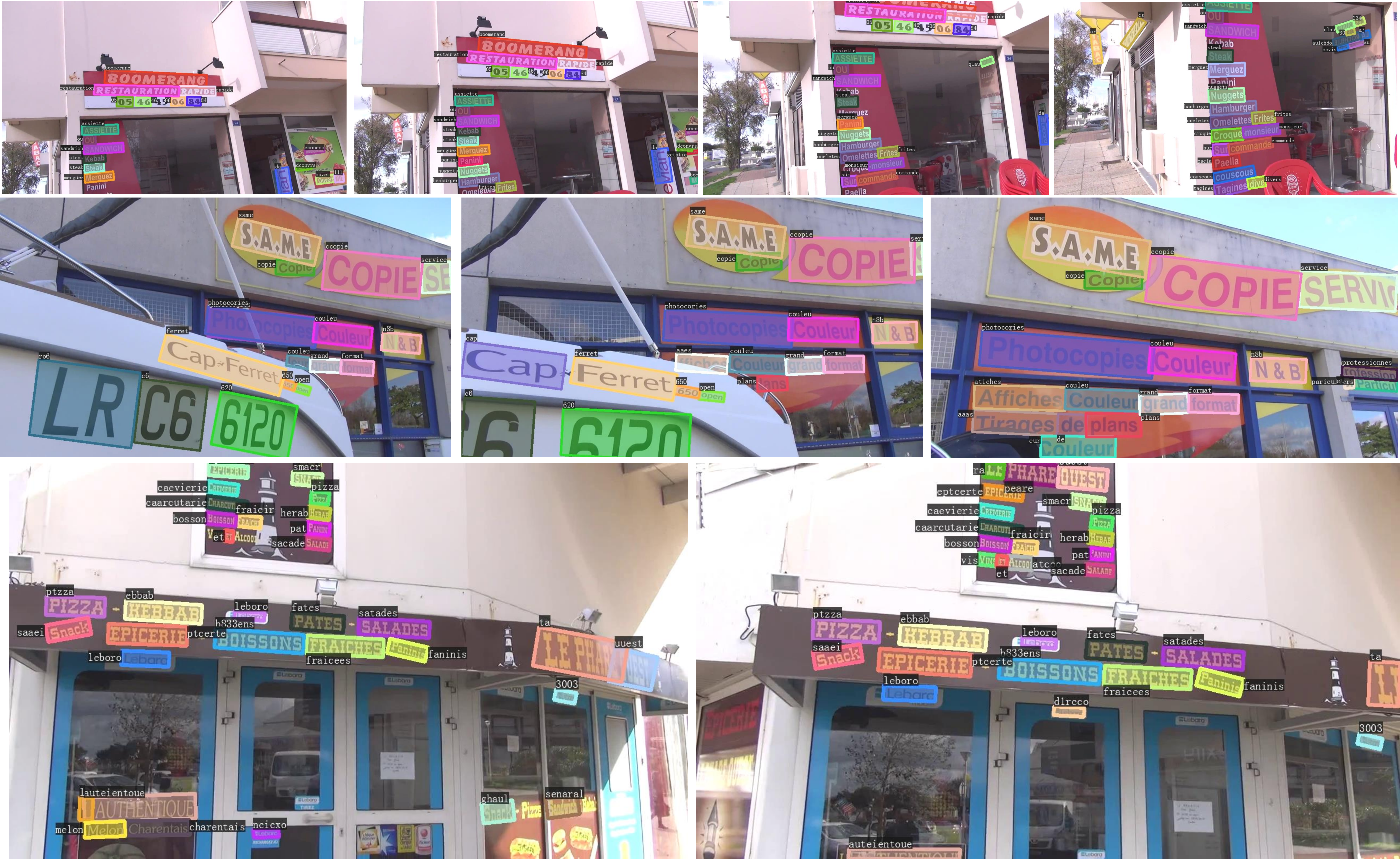}
\caption{\textbf{Visualization of \detr for video text spotting results on ICDAR2015video~\cite{karatzas2015icdar}.} }
\label{Visualization2}
\end{center}
\end{figure*}

\begin{table*}[h]
    \centering
    \setlength{\tabcolsep}{1mm}
    \caption{\textbf{Video text spotting performance on ICDAR2015(video).} `aug' denotes data augmentation for random cropping.}
    \label{tab:icdar15}
\footnotesize 
\begin{tabular}{l | c | ccccc | c}

\multirow{2}{*}{Method} &\multirow{2}{*}{Backbone} & \multicolumn{5}{c|}{End to End Video Text Spotting/\%} & \multirow{2}{*}{FPS}\\
\cline{3-7} 
&& ${\rm ID_{F1}}$$\uparrow$  & ${\rm MOTA}$$\uparrow$ & ${\rm MOTP}$$\uparrow$ & ${\rm M\mbox{-}Tracked}$$\uparrow$ & ${\rm M\mbox{-}Lost}$$\downarrow$ & \\
\shline \hline
\textit{Separate Framework}& & & & && \\
USTB\_TexVideo(II\mbox{-}2)~\cite{karatzas2015icdar} & & 21.3 & 13.2 &  66.6 & 6.6& 67.7 & - \\
USTB\_TexVideo~\cite{karatzas2015icdar} & & 28.2 & 15.6 & 68.5 & 9.5& 60.7& -\\
StradVision\mbox{-}1~\cite{karatzas2015icdar} & & 32.0 & 9.0 &  70.2 & 8.9& 59.5 & -\\
Free~\cite{cheng2020free} & ResNet50 & 61.9 & 53.0 &  74.9 & \textbf{45.5} & 35.9 & 8.8\\
 TransVTSpotter\cite{wu2021bilingual} & ResNet50 & 61.5 & 53.2 &  74.9 & - & - & 9.0\\
\hline
\textit{End-to-End framework}& & & & & && \\
\detr  & ResNet50 & 70.4 & 58.4 & \textbf{75.2} & 32.0 & 20.8 & 16.7 \\
\detr(aug)  & ResNet50 & \textbf{72.8} & \textbf{60.9} & 74.6 & 33.6 & \textbf{20.8} & \textbf{16.7} \\

\end{tabular}
\end{table*}

\textbf{Video text tracking.} Since the tracking task is the core task in this work, we adopt four common datasets(\ie{} ICDAR2015~\cite{karatzas2015icdar}, YVT~(video)~\cite{nguyen2014video}, ICDAR2013~(video)~\cite{karatzas2013icdar} and Minetto~\cite{minetto2011snoopertrack}) to evaluate the task.
Table~\ref{tab:icdar13} presents performance comparison on ICDAR2015~(video)~\cite{karatzas2015icdar}.
Without convolutions and hand-crafted components such as NMS, our \detr achieves a great improvement of $7.6\%$ ${\rm ID_{F1}}$ than the current SOTA method~\cite{cheng2020free}.
This is reasonable, ${\rm M\mbox{-}Tracked}$ of \detr is obvious improvement with $5.4\%$, which denotes \detr can track text instance more robustly and accurately.
For the improvement, Fig.~\ref{Visualization} presents more direct visualization with less false negatives.
Besides, compare with the previous works(\eg{} Free~\cite{cheng2020free}) using multi-stage matching and refinement, our \detr is a simple end-to-end framework.
As for ICDAR2013 video dataset, compared with YORO~\cite{cheng2019you}, our \detr present better performance with $4.7\%$ ${\rm ID_{F1}}$ and $7.4\%$ ${\rm MOTA}$ improvements, while its inference speed is faster.
For Minetto and YVT datasets, our \detr also shows a powerful performance with $2.4\%$ and $1.7\%$ MOTA improvements than the previous methods, respectively.
The better performance on YVT verify the robustness and generalization of our \detr, while the model is trained on ICDAR2015 video and test directly on its test set.
Besides, to further present the effectiveness of \detr, we visualize the tracking and spotting results of Free~\cite{cheng2020free}(SOTA method) and \detr in Figure~\ref{Visualization}.
Compared with Free~\cite{cheng2020free}, our \detr shows a more stable and accurate tracking trajectory by alleviating the problem of ID switching and object missing~(red circle).
And more statistics and analysis can be found in Fig.~\ref{Visualization3} and ablation study.

\textbf{Video text spotting.}
Video text spotting task, as the final aim, requires one to simultaneously track and recognize text instances accurately in a video sequence.
Table~\ref{tab:icdar15} provides the performance comparison of the task on ICDAR2015(video)~\cite{karatzas2015icdar}.
\detr inherits the advantage from query-based tracking, achieves $11.3\%$ (${\rm ID_{F1}}$) with $7.7\%$(${\rm MOTA}$) improvements than the current SOTA method, while the inference speed is faster~($16.7$ FPS $v.s$ $9.0$ FPS).
Meanwhile, \detr drastically reduces the `Mostly Lost' by alleviating up to $15.0\%$ false negatives. 
Besides, comparing the SOTA method Free~\cite{cheng2020free} with a complicated pipeline(\eg{} NMS, Aggregation, Match and Score), our \detr is a simple and clean framework without hand-crafted components.
And more surprising is our \detr is a real end-to-end trainable framework, compared previous works with multiple models and hand-crafted strategies.
To further present the effectiveness of the proposed method, we list more high-quality visualization results in Figure.~\ref{Visualization2}.

\section{Potential Negative Societal Impact and Limitations}
\textbf{Potential Negative Societal Impact}
Similar to the previous video text spotting methods~\cite{cheng2020free}, the proposed \detr have a potential negative risk.
Some important private information in video can be easily collected automatically by the method engaged in fraud, marketing, or other illegal activities.

\textbf{Limitations}
The proposed method shows great performance on video text detection, tracking, and spotting tasks. 
But the speech with around $10$ FPS is a limitation for its application.
Compared with the existing real-time image-based detector, the speech of our \detr still needs further optimization.
And we will continue to explore a real-time transformer-based video text spotting method in future work.

\section{Conclusion}
In this paper, we firstly present an end-to-end trainable video text \textbf{D}Etection, \textbf{T}racking and \textbf{R}ecognition framework with Transformers~(\detr), which views the VTS task as a direct long-range temporal variation problem.
Without explicit heuristics, \detr models each temporal arbitrary-oriented boxes tracking trajectory and text content implicitly by different `text query'. 
With temporal tracking loss over multiple frames and recognition head, the overall pipeline is significantly different and simpler than existing approaches.
%
Without convolutions and hand-crafted components such as dense anchors and NMS, our \detr achieves $65.5\%$ ${\rm ID_{F1}}$ for tracking task and $72.8\%$ ${\rm ID_{F1}}$ for video text spotting task on ICDAR2015~(video), with $8.2\%$ and $11.3\%$ improvements than the previous SOTA method, respectively.
Meantime, the inference time is faster than previous methods.
We hope that the long-range modeling approach can be applied to more video-and-language tasks in the future.

{\small
\bibliographystyle{ieee_fullname}
\bibliography{arxiv}
}
\end{document}